\pdfoutput=1

\documentclass[11pt]{article}
\usepackage[table]{xcolor}
\usepackage{tabularx}
\usepackage{lscape}

\usepackage{acl}

\usepackage{times}
\usepackage{latexsym}

\usepackage{graphicx}
\usepackage{enumitem}
\usepackage{multirow}
\usepackage{caption}
\usepackage{subcaption}
\usepackage{booktabs}

\usepackage[T1]{fontenc}

\usepackage[utf8]{inputenc}

\usepackage{microtype}

\newif\ifshowcomments
 \showcommentsfalse

\title{NewsQs: Multi-Source Question Generation for the Inquiring Mind}

\author{Alyssa Hwang\textsuperscript{2}\thanks{\hspace{1.5mm}Work conducted during an internship at Amazon.}\hspace{1mm} \hspace{5mm} Kalpit Dixit\textsuperscript{1}\thanks{\hspace{1.5mm}Corresponding Author} \hspace{5mm} Miguel Ballesteros\textsuperscript{1} \\ {\bf Yassine Benajiba\textsuperscript{1} \hspace{5mm} Vittorio Castelli\textsuperscript{1} \hspace{5mm} Markus Dreyer\textsuperscript{3}} \\ {\bf Mohit Bansal\textsuperscript{1} \hspace{3mm} Kathleen McKeown\textsuperscript{1}}\\
\textsuperscript{1}AWS AI Labs \hspace{3mm} \textsuperscript{2}University of Pennsylvania \hspace{3mm}
\textsuperscript{3}Alexa \\
\texttt{ahwang16@seas.upenn.edu} \\
\texttt{\{kddixit, ballemig, benajiy, vittorca, }\\
\texttt{mddreyer, mobansal, mckeownk\}@amazon.com} \\
}

\begin{document}
\maketitle
\begin{abstract}
We present NewsQs (\textit{news-cues}), a dataset that provides question-answer pairs for multiple news documents. To create NewsQs, we augment a traditional multi-document summarization dataset with questions automatically generated by a T5-Large model fine-tuned on FAQ-style news articles from the News On the Web corpus. We show that fine-tuning a model with control codes produces questions that are judged acceptable more often than the same model without them as measured through human evaluation. We use a QNLI model with high correlation with human annotations to filter our data. We release our final dataset of high-quality questions, answers, and document clusters as a resource for future work in query-based multi-document summarization.\footnote{To be released upon publication.}
\end{abstract}

\section{Introduction}

Providing the ability to answer questions about events and people in the news would help compensate for information overload in the modern digital age.
Curating datasets that teach machines to answer such questions, however, is challenging because the datasets require three components: clusters of documents each conveying different aspects of an event or person, questions covering the documents, and answers to the questions.
Existing datasets related to this area draw source material from social media, Wikipedia, or stories, leaving a gap for news-based resources.

\begin{figure}[h]
    \centering
    \includegraphics[width=\columnwidth]{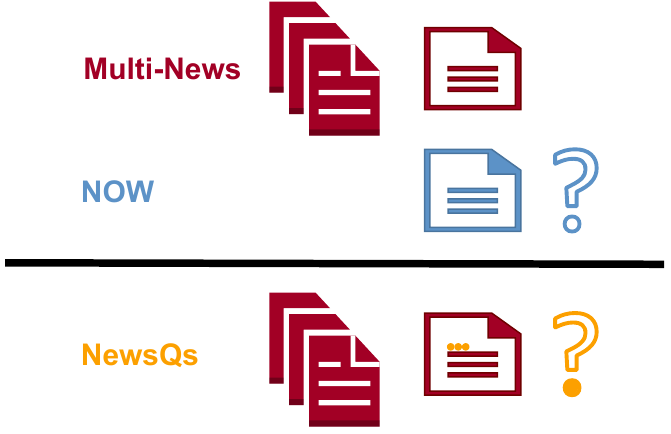}
    \caption{Resources used to create NewsQs. Multi-News contains document clusters and summaries. QA pairs from NOW are used for fine-tuning.}
    \label{fig:overview}
\end{figure}

We generate our dataset, NewsQs (\textit{news-cues}), by augmenting an existing dataset called Multi-News \cite{fabbri_multi-news_2019} 
with automatically generated questions. Multi-News is a multi-document summarization (MDS) dataset that provides clusters of documents and human-written summaries. Augmenting Multi-News with questions would enable NewsQs to serve as a resource for \textit{query-based} multi-document summarization (qMDS), which is the task of generating paragraph-long answers to open-ended questions by drawing information from multiple sources. Pivoting our dataset generation task into a question generation (QG) task also simplifies the process because we no longer need to collect all three necessary components. We already have the documents and answers; we just need to generate the questions to connect them.

In our work, we experiment with fine-tuning a large language model on FAQ-style news articles from the News On the Web (NOW) corpus \cite{davies_corpus_2022} to generate questions for Multi-News. Our experiments show that adding control codes while fine-tuning helps the model generate topical questions related to the entire summary. Since we do not have reference questions for automatic evaluation, we use a QNLI model with high correlation with human annotations to discard low-quality examples.
Our contributions include:

\begin{itemize}
    \itemsep0em
    \item A dataset, NewsQs, of 21,000 high-quality question-answer pairs for multiple documents.
    \item A method for using two existing datasets, each containing two of the three necessary components, to create NewsQs.
    \item A human evaluation task designed for long text that demonstrates fine-tuning with control codes produces better quality questions.
\end{itemize}

\section{Related Work}
Datasets for query-based multi-document summarization (qMDS) need (1) multiple source documents, (2) questions, and (3) long-form answers.
Only a few datasets fulfill all three requirements: ELI5 from KILT \cite{petroni-etal-2021-kilt}, ASQA \cite{stelmakh_asqa_2022}, DuReader \cite{he_dureader_2018}, AQuaMuSe  \citep{kulkarni_aquamuse_2020}, and SQuALITY \cite{wang_squality_2022}. Concerningly, these datasets contain source documents that were automatically gathered by a variety of heuristics, so the original ELI5 dataset provides support documents that contain the full answer only 65\% of the time \citep{fan_eli5_2019}. Providing source documents that do not fully contain the answer teaches models to hallucinate \cite{krishna-etal-2021-hurdles}.

Other related resources for qMDS are missing at least one component. Multi-document summarization datasets contain multiple source documents and summaries but not questions \cite{duc_document_2007, gholipour_ghalandari_large-scale_2020}. Multi-document question answering datasets contain short answers of up to a few words rather than longer, more detailed responses. Through question generation, we are able to leverage these existing resources to construct a dataset with all three components that is substantially larger \citep{duan_question_2017, dugan-etal-2022-feasibility, chakrabarty_consistent_2022}.

\section{Methods}
Our goal is to produce a dataset of questions and answers that are relevant to topical document clusters.
Our approach leverages two existing datasets, each providing only two of the three components.
We experiment with three methods for a range of models (see Appendix~\ref{sec:model_specific_prompts} for prompts).

\subsection{Data}
A large dataset containing two of the three necessary components already exists: Multi-News \citep{fabbri_multi-news_2019}. This dataset contains approximately 56,000 clusters of related news articles and human-written summaries. Zero-shot generation of questions on Multi-News does not produce suitable questions (see Section~\ref{sec:auto_eval}), so we experiment with fine-tuning models on a small training set of similar data from the News On the Web (NOW) corpus \citep{davies_corpus_2022}.

We sample 1,200 question-answer pairs from FAQ-style articles---a popular style of news article that presents information as a series of Frequently Asked Questions---and split them into 80/20 train/validation sets. We manually exclude QA pairs that are not self-contained, such as a question like ``What about it?'' We also exclude QA pairs with answers that are fewer than 30 words or longer than 350 words to keep our fine-tuning dataset similar to our final inference dataset; Multi-News summaries have an average of 266 words. An example question-answer pair from NOW is shown in Appendix Table~\ref{tab:now_examples}.

\subsection{Fine-Tuning (FT) Experiments}
\label{sec:finetuning}

Based on preliminary zero-shot experiments, we fine-tune T5-Large to generate questions for NewsQs (see Appendix Table~\ref{tab:mn_examples}).

\paragraph{Vanilla} We fine-tune T5-Large by including the paragraph-long NOW answer followed by the reference question and training for 100, 200, and 500 epochs with learning ranges of $1.00 \times 10^{-4}$, $3.00 \times 10^{-4}$, and $1.00 \times 10^{-3}$, for a total of 9 experiments. The best settings in our parameter search were 500 epochs with a learning rate of $3.00 \times 10^{-4}$, which we consider our baseline.

\paragraph{Control Codes} \label{sec:control_codes_fine_tuning} We experiment with prepending prompts with control codes (i.e., [dogs; cats; pets]) before fine-tuning with the baseline settings to encourage T5-Large to write a question about the full paragraph (see Appendix~\ref{sec:t5_large_control_code_prompt}). We use salient keywords or entities as control codes because main ideas of news articles tend to come from these types of words. After investigating different numbers of output from Yet Another Keyword Extractor \citep{campos_yake_2020, campos_yake_2018, pasi_yake_2018} and an off-the-shelf entity detection model, we use the top three outputs from the entity detection model as the control codes.

\section{Results of Fine-Tuning Experiments}
\label{sec:auto_eval}
We show performance on the NOW training set using ROUGE-L \cite{lin-2004-rouge} and BERTScore \cite{zhang_bertscore_2019} F1-scores (see Table~\ref{tab:t5-large}).
The improvement from zero-shot T5-Large to fine-tuning with T5-Large is clearly evident;   ROUGE-L F1-score increases by 0.2 absolute and BERTScore by 0.42. Automatic metrics do not show an improvement when control codes are added. A spot check of questions by the authors, however, suggested  that generated questions are more appropriate for the full paragraph answer (see examples in Appendix Table~\ref{tab:mn_examples}), so we continued with human evaluation.

\begin{table}[]
\centering
\begin{tabular}{@{}lcc@{}}
\toprule
                 & ROUGE-L & BERTScore \\ \midrule
Vanilla ZS      & 0.12       & -0.05        \\ 
Vanilla FT      & 0.38       & 0.47         \\ 
Control Codes FT & 0.39       & 0.47         \\ \bottomrule
\end{tabular}
\caption{Fine-tuning results. Automatic metrics show improvement after fine-tuning but not after adding control codes, which is more apparent in human evaluation.}
\label{tab:t5-large}
\end{table}

\begin{table}[]
\centering
\begin{tabular}{@{}lccc@{}}
\toprule
 & Train & Val. & Test \\ \midrule
Number of Examples & 17K & 2.1K & 1.6K \\
Avg. Len. of Question     & 11.0 & 11.0 & 11.1 \\
Avg. Len. of Answer & 288 & 288 & 287 \\
\# Entities Per Answer      & 8.46 & 8.55 & 8.55  \\
\% Overlap of Entities & 43.2 & 43.4 & 44.1 \\ 
Avg. QNLI Score & \textbf{0.936} & \textbf{0.939} & \textbf{0.941} \\
\midrule
Avg. QNLI Score (BF) & 0.387 & 0.387 & 0.396 \\
\bottomrule
\end{tabular}
\caption{NewsQs dataset statistics. QNLI scores substantially improved before (BF) and after filtering. The lengths of questions and answers are balanced across splits. The percentage overlap of entities between questions and answers indicates that generated questions cover a fair amount of the main ideas in the answers.}
\label{tab:dataset_statistics}
\end{table}

\begin{table}[]
\centering
\begin{tabular}{@{}lccc@{}}
\toprule
 & Train & Val. & Test  \\ \midrule
who           & 6.77  & 7.15       & 6.32  \\
what          & 48.2 & 46.7     &  49.6 \\
when          & 1.50  & 1.34       & 1.06  \\
where         & 2.51  & 2.54       & 2.70  \\
why           & 8.70  & 9.32       & 7.82  \\
how           & 22.8 & 22.8      & 23.1 \\ \bottomrule
\end{tabular}
\caption{Percentage of questions in each data split that start with the specified question words.}
\label{tab:app_question_types}
\end{table}

\section{Human Evaluation}

\begin{table*}[]
\centering
\begin{tabular}{lcccc||cccc}
\toprule
  & \multicolumn{2}{c}{Sentence Relevance ($\kappa$)} & \multicolumn{2}{c||}{Question Acceptability ($\kappa$)} & \multicolumn{4}{c}{Ranking (\%)}                                                      \\
  & All             & \cellcolor[HTML]{ECF4FF} Accept. Only             & Multiclass          & \cellcolor[HTML]{ECF4FF} Binary         & Mean  & >                    & <                 & =                   \\
  \midrule
Vanilla & 0.32             & \cellcolor[HTML]{ECF4FF} 0.32                & 0.20                & \cellcolor[HTML]{ECF4FF} 0.27           & 2.32 & \multirow{2}{*}{45.7} & \multirow{2}{*}{27.2} & \multirow{2}{*}{27.1} \\
Control Codes & 0.24             & \cellcolor[HTML]{ECF4FF} 0.56                & 0.26                & \cellcolor[HTML]{ECF4FF} 0.34           & 2.53 &                         &                        &                       
\\ \bottomrule
\end{tabular}
\caption{Results of human evaluation for both fine-tuned models. We report Cohen's Kappa (left) for all and acceptable questions. We also compute scores for each question by ranking the question labels. We report the mean for each model and the percentage of times the control codes model had a higher, lower, or equal score (right).}
\label{tab:human_eval_analysis}
\end{table*}

Since we ran inference on the Multi-News dataset to create new questions, we do not have reference questions for automatic metrics. 
We need human input to evaluate the final output of our model.

\subsection{Design}

Evaluating our data requires human annotators to read questions and paragraph-long answers, which can be very cognitively demanding. We designed a human evaluation task that breaks down long text into sentence-level annotations to make this process easier. The task is composed of two subtasks: Sentence Relevance and Question Acceptability.

For each question in NewsQs, we show the answer one sentence at a time and ask annotators to judge if each sentence is relevant to the question. When 
they have seen the entire paragraph, annotators are asked to rate the question as one of the following: acceptable, narrow topic, broad topic, irrelevant topic, wrong type, incoherent, incomplete, not a question, or other. Along with decreasing the amount of new information annotators have to review at a time, our task collects human judgment at the sentence and paragraph levels.

\subsection{Results and Discussion}
At least one annotator judged 81.6\% of control codes questions as acceptable, demonstrating the high quality of our questions.
We report Cohen's Kappa scores across tasks and experiments in the left half of Table~\ref{tab:human_eval_analysis}. Two annotators evaluated data for each experiment, for a total of four annotators in our study.
When inspecting agreement for each input paragraph, we realized that inter-annotator agreement on Sentence Relevance was much lower for questions marked as not acceptable because relevance to a low-quality question is not well defined. 
When including only questions marked as acceptable by at least one annotator, inter-annotator agreement increases substantially: from a Cohen's Kappa of 0.32 (fair) to 0.56 (moderate) for Sentence Relevance on the control codes FT model.

We estimated the level of quality of a question by ranking the possible question judgments. A judgment of acceptable was assigned rank 3, broad or narrow topic was assigned rank 2, and the remaining  were assigned rank 1. 
We then computed an average score for all 188 questions.
The average vanilla model ranking was 2.32 and the average control codes model ranking was 2.53, where the rankings ranged from 1 to 3 and higher is better. 
These results are significant with p-value 0.0002, according to the Bootstrap Test \cite{dror_hitchhikers_2018}.
For a deeper analysis, we compared all of the control codes and vanilla scores.
As shown on the right side of Table~\ref{tab:human_eval_analysis}, the control codes model scores higher 45\% of the time and equal 27.1\%. These two methods for analyzing human evaluation results show that the control codes FT model outperforms the vanilla FT model on acceptability.

By also asking annotators to rate relevance one sentence at a time, we were able to collect finer-grained data on how relevant a question is to an input paragraph. We computed the Sentence Relevance Score (SRS) for each question by counting the number of times both annotators judged a sentence as relevant to the question and normalizing by the number of sentences. SRS ranges from 0 to 1, where higher is better. The average SRS of the vanilla and control codes models was 0.86 and 0.96, respectively.
These results demonstrate that questions generated with control codes are more likely to require a larger portion of the sentences in the summary for a suitable response.

\section{Automatic Evaluation}

Since we generated new questions to augment Multi-News, we do not have reference questions for automatic metrics like ROUGE or BERTScore. We instead use a QNLI model to measure if the question is answerable given the paragraph. We find that QNLI correlates highly with human judgments of question acceptability ($\rho = 0.41$) and passage relevance ($\rho = 0.23$). We use QNLI to filter our dataset from 54K to 21K high-quality examples (17,377 train/2,167 validation/1,597 test).

\begin{table}[!h]
    \centering
    \begin{tabular}{>{\raggedright}p{7cm}}
        \toprule
        NewsQs Examples \tabularnewline
        \midrule
        \textbf{Question} What's the most earth-like planet in the solar system? \newline
        \textbf{Answer} if you're sick of earth and have a spacecraft capable of traveling hundreds of light-years, astronomers have spotted the most promising destination yet. kepler-438b is a newly-confirmed potential ``earth twin'' detected by the kepler space telescope, the bbc reports, one of eight confirmed new exoplanets. [+174 words]
 \newline
        \textbf{Documents} [2,949 words from 3 articles]
        \tabularnewline
        \midrule

        \textbf{Question} Do ants really just turn left? \newline
        \textbf{Answer} roughly nine in 10 humans are right-handed, an example of ``brain lateralization'' that's pretty common among vertebrates—and now apparently invertebrates. researchers in the uk are finding that even ants—which are invertebrates, meaning they have exoskeletons—carry an innate directional bias, in their case almost always turning left when exploring new territory, reports science daily. [+94 words] \newline
        \textbf{Documents} [486 words from 2 articles]
        \tabularnewline
        \bottomrule
    \end{tabular}
    \caption{Examples from our NewsQs dataset. Each example in NewsQs contains a machine-generated question and an answer and cluster of source documents from Multi-News.}
    \label{tab:newsqs_examples}
\end{table}

\section{Discussion and Conclusion}

In our work, we presented a method for fine-tuning T5-Large with control codes to generate questions that can be answered by an input paragraph. We used this method to create NewsQs and showed that it produces acceptable questions 80.3\% of the time, which were further filtered with a QNLI model.
Designing a good human evaluation task for NewsQs was challenging because the natural instinct to display the question with the input paragraph would have been very cognitively demanding. In discussions with the team that evaluated NewsQs, the annotators asked us to ``remember the human in human evaluation,'' so we adjusted by displaying paragraphs one sentence at a time. We believe that more work in human-centered annotation tasks will improve evaluation. Moreover, remembering the human will improve NLP as a whole, for annotators, researchers, and users alike.

\section*{Limitations}
Our work augments an existing dataset, Multi-News, which means that it depends on its quality and inherits its limitations. We chose the Multi-News dataset because it is publicly available, vetted and published by a reputable venue, and curated through careful human effort. Humans, however, can still introduce error even in the most carefully designed settings. Despite using a qualified team of linguistics experts for human evaluation, inter-annotator agreement is low, suggesting that the complexity of NLP tasks makes high agreement difficult to achieve. 

We cannot release our fine-tuned models because our fine-tuning dataset, the News On the Web (NOW) corpus, is not open to the public. The NOW corpus, however, generously offers free access to researchers affiliated with an educational institution who apply for it. We use a very small subset of the articles contained in the NOW corpus, which were scraped from publicly available websites. Researchers who are interested in reproducing our work or generating similar questions on a different inference set can use our methods, which were designed for modest computational resources---our fine-tuning method takes about two days per T5-Large model on an AWS p3.16x. This also comes with an important trade-off: having smaller resources means using smaller models, and smaller models are known to perform worse than newer, much larger models. We believe that our work helps combat this trade-off, allowing more people to use NLP technology.

\section*{Ethical Considerations}
We used email to recruit a qualified team of linguistics experts to evaluate our dataset, ensuring that payment for their services was fair given their country of residence and amount of work the task required. The experts are based in the United States and diverse in gender, ethnicity, country of origin, and languages spoken. We provided in-depth instructions on how their annotations would be used and hosted several meetings throughout the process to introduce our task, answer questions, and improve our evaluation design.

Our work presents a dataset for explaining multiple news articles, which vary widely in content. We manually inspected our fine-tuning dataset from the NOW corpus for quality and ensured that information about the author of each article was not included during training. We add machine-generated questions to the Multi-News dataset, which has already been vetted and published, without modifying it. To the best of our knowledge through periodic empirical analysis and final human evaluation, the questions we generate by fine-tuning on NOW and running inference on Multi-News have low risk of causing harm. We will release the questions judged acceptable by both annotators as a gold test set upon publication for future work.



\bibliography{anthology,custom,references}

\begin{thebibliography}{24}
\expandafter\ifx\csname natexlab\endcsname\relax\def\natexlab#1{#1}\fi

\bibitem[{Campos et~al.(2020)Campos, Mangaravite, Pasquali, Jorge, Nunes, and
  Jatowt}]{campos_yake_2020}
Ricardo Campos, Vítor Mangaravite, Arian Pasquali, Alípio Jorge, Célia
  Nunes, and Adam Jatowt. 2020.
\newblock {YAKE}! {Keyword} extraction from single documents using multiple
  local features.
\newblock \emph{Information Sciences}, 509:257--289.
\newblock Publisher: Elsevier.

\bibitem[{Campos et~al.(2018{\natexlab{a}})Campos, Mangaravite, Pasquali,
  Jorge, Nunes, and Jatowt}]{campos_yake_2018}
Ricardo Campos, Vítor Mangaravite, Arian Pasquali, Alípio~Mário Jorge,
  Célia Nunes, and Adam Jatowt. 2018{\natexlab{a}}.
\newblock Yake! collection-independent automatic keyword extractor.
\newblock In \emph{European {Conference} on {Information} {Retrieval}}, pages
  806--810. Springer.

\bibitem[{Campos et~al.(2018{\natexlab{b}})Campos, Mangaravite, Pasquali,
  Jorge, Nunes, and Jatowt}]{pasi_yake_2018}
Ricardo Campos, Vítor Mangaravite, Arian Pasquali, Alípio~Mário Jorge,
  Célia Nunes, and Adam Jatowt. 2018{\natexlab{b}}.
\newblock \href {https://doi.org/10.1007/978-3-319-76941-7_80} {{YAKE}!
  {Collection}-{Independent} {Automatic} {Keyword} {Extractor}}.
\newblock In Gabriella Pasi, Benjamin Piwowarski, Leif Azzopardi, and Allan
  Hanbury, editors, \emph{Advances in {Information} {Retrieval}}, volume 10772,
  pages 806--810. Springer International Publishing, Cham.
\newblock Series Title: Lecture Notes in Computer Science.

\bibitem[{Chakrabarty et~al.(2022)Chakrabarty, Lewis, and
  Muresan}]{chakrabarty_consistent_2022}
Tuhin Chakrabarty, Justin Lewis, and Smaranda Muresan. 2022.
\newblock \href {https://doi.org/10.48550/ARXIV.2210.11536} {{CONSISTENT}:
  {Open}-{Ended} {Question} {Generation} {From} {News} {Articles}}.

\bibitem[{Davies(2022)}]{davies_corpus_2022}
Mark Davies. 2022.
\newblock \href {https://www.english-corpora.org//now/} {Corpus of {News} on
  the {Web} ({NOW})}.

\bibitem[{Dror et~al.(2018)Dror, Baumer, Shlomov, and
  Reichart}]{dror_hitchhikers_2018}
Rotem Dror, Gili Baumer, Segev Shlomov, and Roi Reichart. 2018.
\newblock \href {http://aclweb.org/anthology/P18-1128} {The {Hitchhiker}'s
  {Guide} to {Testing} {Statistical} {Significance} in {Natural} {Language}
  {Processing}}.
\newblock In \emph{Proceedings of the 56th {Annual} {Meeting} of the
  {Association} for {Computational} {Linguistics} ({Volume} 1: {Long}
  {Papers})}, pages 1383--1392. Association for Computational Linguistics.
\newblock Event-place: Melbourne, Australia.

\bibitem[{Duan et~al.(2017)Duan, Tang, Chen, and Zhou}]{duan_question_2017}
Nan Duan, Duyu Tang, Peng Chen, and Ming Zhou. 2017.
\newblock \href {https://doi.org/10.18653/v1/D17-1090} {Question {Generation}
  for {Question} {Answering}}.
\newblock In \emph{Proceedings of the 2017 {Conference} on {Empirical}
  {Methods} in {Natural} {Language} {Processing}}, pages 866--874, Copenhagen,
  Denmark. Association for Computational Linguistics.

\bibitem[{{DUC}(2007)}]{duc_document_2007}
{DUC}. 2007.
\newblock \href {https://duc.nist.gov/data.html} {Document {Understanding}
  {Conferences} - {Past} {Data}}.

\bibitem[{Dugan et~al.(2022)Dugan, Miltsakaki, Upadhyay, Ginsberg, Gonzalez,
  Choi, Yuan, and Callison-Burch}]{dugan-etal-2022-feasibility}
Liam Dugan, Eleni Miltsakaki, Shriyash Upadhyay, Etan Ginsberg, Hannah
  Gonzalez, DaHyeon Choi, Chuning Yuan, and Chris Callison-Burch. 2022.
\newblock \href {https://doi.org/10.18653/v1/2022.findings-acl.151} {A
  feasibility study of answer-agnostic question generation for education}.
\newblock In \emph{Findings of the Association for Computational Linguistics:
  ACL 2022}, pages 1919--1926, Dublin, Ireland. Association for Computational
  Linguistics.

\bibitem[{Fabbri et~al.(2019)Fabbri, Li, She, Li, and
  Radev}]{fabbri_multi-news_2019}
Alexander Fabbri, Irene Li, Tianwei She, Suyi Li, and Dragomir Radev. 2019.
\newblock \href {https://doi.org/10.18653/v1/P19-1102} {Multi-{News}: {A}
  {Large}-{Scale} {Multi}-{Document} {Summarization} {Dataset} and
  {Abstractive} {Hierarchical} {Model}}.
\newblock In \emph{Proceedings of the 57th {Annual} {Meeting} of the
  {Association} for {Computational} {Linguistics}}, pages 1074--1084, Florence,
  Italy. Association for Computational Linguistics.

\bibitem[{Fan et~al.(2019)Fan, Jernite, Perez, Grangier, Weston, and
  Auli}]{fan_eli5_2019}
Angela Fan, Yacine Jernite, Ethan Perez, David Grangier, Jason Weston, and
  Michael Auli. 2019.
\newblock \href {http://arxiv.org/abs/1907.09190} {{ELI5}: {Long} {Form}
  {Question} {Answering}}.
\newblock \emph{arXiv:1907.09190 [cs]}.
\newblock ArXiv: 1907.09190.

\bibitem[{Gholipour~Ghalandari et~al.(2020)Gholipour~Ghalandari, Hokamp, Pham,
  Glover, and Ifrim}]{gholipour_ghalandari_large-scale_2020}
Demian Gholipour~Ghalandari, Chris Hokamp, Nghia~The Pham, John Glover, and
  Georgiana Ifrim. 2020.
\newblock \href {https://www.aclweb.org/anthology/2020.acl-main.120} {A
  {Large}-{Scale} {Multi}-{Document} {Summarization} {Dataset} from the
  {Wikipedia} {Current} {Events} {Portal}}.
\newblock In \emph{Proceedings of the 58th {Annual} {Meeting} of the
  {Association} for {Computational} {Linguistics}}, pages 1302--1308, Online.
  Association for Computational Linguistics.

\bibitem[{He et~al.(2018)He, Liu, Liu, Lyu, Zhao, Xiao, Liu, Wang, Wu, She,
  Liu, Wu, and Wang}]{he_dureader_2018}
Wei He, Kai Liu, Jing Liu, Yajuan Lyu, Shiqi Zhao, Xinyan Xiao, Yuan Liu,
  Yizhong Wang, Hua Wu, Qiaoqiao She, Xuan Liu, Tian Wu, and Haifeng Wang.
  2018.
\newblock \href {https://doi.org/10.48550/arXiv.1711.05073} {{DuReader}: a
  {Chinese} {Machine} {Reading} {Comprehension} {Dataset} from {Real}-world
  {Applications}}.
\newblock ArXiv:1711.05073 [cs].

\bibitem[{Krishna et~al.(2021)Krishna, Roy, and
  Iyyer}]{krishna-etal-2021-hurdles}
Kalpesh Krishna, Aurko Roy, and Mohit Iyyer. 2021.
\newblock \href {https://doi.org/10.18653/v1/2021.naacl-main.393} {Hurdles to
  progress in long-form question answering}.
\newblock In \emph{Proceedings of the 2021 Conference of the North American
  Chapter of the Association for Computational Linguistics: Human Language
  Technologies}, pages 4940--4957, Online. Association for Computational
  Linguistics.

\bibitem[{Kulkarni et~al.(2020)Kulkarni, Chammas, Zhu, Sha, and
  Ie}]{kulkarni_aquamuse_2020}
Sayali Kulkarni, Sheide Chammas, Wan Zhu, Fei Sha, and Eugene Ie. 2020.
\newblock \href {https://doi.org/10.48550/arXiv.2010.12694} {{AQuaMuSe}:
  {Automatically} {Generating} {Datasets} for {Query}-{Based}
  {Multi}-{Document} {Summarization}}.

\bibitem[{Lewis et~al.(2019)Lewis, Liu, Goyal, Ghazvininejad, Mohamed, Levy,
  Stoyanov, and Zettlemoyer}]{lewis_bart_2019}
Mike Lewis, Yinhan Liu, Naman Goyal, Marjan Ghazvininejad, Abdelrahman Mohamed,
  Omer Levy, Ves Stoyanov, and Luke Zettlemoyer. 2019.
\newblock \href {https://doi.org/10.48550/arXiv.1910.13461} {{BART}:
  {Denoising} {Sequence}-to-{Sequence} {Pre}-training for {Natural} {Language}
  {Generation}, {Translation}, and {Comprehension}}.

\bibitem[{Lin(2004)}]{lin-2004-rouge}
Chin-Yew Lin. 2004.
\newblock \href {https://aclanthology.org/W04-1013} {{ROUGE}: A package for
  automatic evaluation of summaries}.
\newblock In \emph{Text Summarization Branches Out}, pages 74--81, Barcelona,
  Spain. Association for Computational Linguistics.

\bibitem[{Petroni et~al.(2021)Petroni, Piktus, Fan, Lewis, Yazdani, De~Cao,
  Thorne, Jernite, Karpukhin, Maillard, Plachouras, Rockt{\"a}schel, and
  Riedel}]{petroni-etal-2021-kilt}
Fabio Petroni, Aleksandra Piktus, Angela Fan, Patrick Lewis, Majid Yazdani,
  Nicola De~Cao, James Thorne, Yacine Jernite, Vladimir Karpukhin, Jean
  Maillard, Vassilis Plachouras, Tim Rockt{\"a}schel, and Sebastian Riedel.
  2021.
\newblock \href {https://doi.org/10.18653/v1/2021.naacl-main.200} {{KILT}: a
  benchmark for knowledge intensive language tasks}.
\newblock In \emph{Proceedings of the 2021 Conference of the North American
  Chapter of the Association for Computational Linguistics: Human Language
  Technologies}, pages 2523--2544, Online. Association for Computational
  Linguistics.

\bibitem[{Radford et~al.(2019)Radford, Wu, Child, Luan, Amodei, and
  Sutskever}]{radford_language_2019}
Alec Radford, Jeffrey Wu, Rewon Child, David Luan, Dario Amodei, and Ilya
  Sutskever. 2019.
\newblock Language {Models} are {Unsupervised} {Multitask} {Learners}.
\newblock page~24.

\bibitem[{Raffel et~al.(2020)Raffel, Shazeer, Roberts, Lee, Narang, Matena,
  Zhou, Li, and Liu}]{raffel_exploring_2020}
Colin Raffel, Noam Shazeer, Adam Roberts, Katherine Lee, Sharan Narang, Michael
  Matena, Yanqi Zhou, Wei Li, and Peter~J. Liu. 2020.
\newblock \href {http://arxiv.org/abs/1910.10683} {Exploring the {Limits} of
  {Transfer} {Learning} with a {Unified} {Text}-to-{Text} {Transformer}}.
\newblock \emph{arXiv:1910.10683 [cs, stat]}.
\newblock ArXiv: 1910.10683.

\bibitem[{Sanh et~al.(2022)Sanh, Webson, Raffel, Bach, Sutawika, Alyafeai,
  Chaffin, Stiegler, Scao, Raja, Dey, Bari, Xu, Thakker, Sharma, Szczechla,
  Kim, Chhablani, Nayak, Datta, Chang, Jiang, Wang, Manica, Shen, Yong, Pandey,
  Bawden, Wang, Neeraj, Rozen, Sharma, Santilli, Fevry, Fries, Teehan, Bers,
  Biderman, Gao, Wolf, and Rush}]{sanh_multitask_2022}
Victor Sanh, Albert Webson, Colin Raffel, Stephen~H. Bach, Lintang Sutawika,
  Zaid Alyafeai, Antoine Chaffin, Arnaud Stiegler, Teven~Le Scao, Arun Raja,
  Manan Dey, M.~Saiful Bari, Canwen Xu, Urmish Thakker, Shanya~Sharma Sharma,
  Eliza Szczechla, Taewoon Kim, Gunjan Chhablani, Nihal Nayak, Debajyoti Datta,
  Jonathan Chang, Mike Tian-Jian Jiang, Han Wang, Matteo Manica, Sheng Shen,
  Zheng~Xin Yong, Harshit Pandey, Rachel Bawden, Thomas Wang, Trishala Neeraj,
  Jos Rozen, Abheesht Sharma, Andrea Santilli, Thibault Fevry, Jason~Alan
  Fries, Ryan Teehan, Tali Bers, Stella Biderman, Leo Gao, Thomas Wolf, and
  Alexander~M. Rush. 2022.
\newblock \href {http://arxiv.org/abs/2110.08207} {Multitask {Prompted}
  {Training} {Enables} {Zero}-{Shot} {Task} {Generalization}}.
\newblock \emph{arXiv:2110.08207 [cs]}.
\newblock ArXiv: 2110.08207.

\bibitem[{Stelmakh et~al.(2022)Stelmakh, Luan, Dhingra, and
  Chang}]{stelmakh_asqa_2022}
Ivan Stelmakh, Yi~Luan, Bhuwan Dhingra, and Ming-Wei Chang. 2022.
\newblock \href {http://arxiv.org/abs/2204.06092} {{ASQA}: {Factoid}
  {Questions} {Meet} {Long}-{Form} {Answers}}.
\newblock \emph{arXiv:2204.06092 [cs]}.
\newblock ArXiv: 2204.06092.

\bibitem[{Wang et~al.(2022)Wang, Pang, Chen, Phang, and
  Bowman}]{wang_squality_2022}
Alex Wang, Richard~Yuanzhe Pang, Angelica Chen, Jason Phang, and Samuel~R.
  Bowman. 2022.
\newblock \href {http://arxiv.org/abs/2205.11465} {{SQuALITY}: {Building} a
  {Long}-{Document} {Summarization} {Dataset} the {Hard} {Way}}.
\newblock \emph{arXiv:2205.11465 [cs]}.
\newblock ArXiv: 2205.11465.

\bibitem[{Zhang et~al.(2019)Zhang, Kishore, Wu, Weinberger, and
  Artzi}]{zhang_bertscore_2019}
Tianyi Zhang, Varsha Kishore, Felix Wu, Kilian~Q. Weinberger, and Yoav Artzi.
  2019.
\newblock \href {https://doi.org/10.48550/arXiv.1904.09675} {{BERTScore}:
  {Evaluating} {Text} {Generation} with {BERT}}.

\end{thebibliography}
\bibliographystyle{acl_natbib}

\appendix

\section{Model-Specific Prompts}
\label{sec:model_specific_prompts}
For T5 models, we start the prompt with a short natural-language command (``generate question'') followed by the answer paragraph and the pad token \citep{raffel_exploring_2020}. Prompts for T0 models are similar, except the short natural-language command is replaced by a sentence-long natural-language instruction (``Give me a question about this answer'') \citep{sanh_multitask_2022}. Prompts for GPT-2 incorporate ``[answer]'' and ``[question]'' special tokens to mark the beginnings of the input answer paragraph and the output question \citep{radford_language_2019}. BART is prompted with the answer paragraph surrounded by beginning-of-sentence and end-of-sentence special tokens \citep{lewis_bart_2019}.

\subsection{T5-Large Control Code Prompt}
\label{sec:t5_large_control_code_prompt}
Following from Section \ref{sec:control_codes_fine_tuning}.
A full prompt with control codes for T5-Large would look like:

\begin{quote}
    generate question: [dogs; cats; pets] Dogs and cats are popular pets\ldots <pad>
\end{quote}

\section{Full Fine-Tuning Results}
We report zero-shot and fine-tuned T5-Large performance across six ngram-, semantic similarity-, and model-based automatic metrics in Table~\ref{tab:full_finetuning}.

\section{Examples}
We show examples from the fine-tuning dataset (News On the Web), inference dataset (Multi-News), and human evaluation of our dataset (NewsQs) in Tables~\ref{tab:now_examples}, \ref{tab:mn_examples}, and \ref{tab:human_eval_examples}.

\begin{landscape}
\begin{table}[]
\centering
\begin{tabular}{@{}llllllllllllll@{}}
\toprule
                 & \multicolumn{3}{c}{ROUGE-1}                                            & \multicolumn{3}{c}{ROUGE-2}                                            & \multicolumn{3}{c}{ROUGE-L}                                            & \multicolumn{3}{c}{BERTScore}                                          & \multicolumn{1}{c}{\multirow{2}{*}{BARTScore}} \\ \cmidrule(r){1-13}
                 & \multicolumn{1}{c}{P} & \multicolumn{1}{c}{R} & \multicolumn{1}{c}{F1} & \multicolumn{1}{c}{P} & \multicolumn{1}{c}{R} & \multicolumn{1}{c}{F1} & \multicolumn{1}{c}{P} & \multicolumn{1}{c}{R} & \multicolumn{1}{c}{F1} & \multicolumn{1}{c}{P} & \multicolumn{1}{c}{R} & \multicolumn{1}{c}{F1} & \multicolumn{1}{c}{}                           \\ \cmidrule(l){14-14} 
Vanilla ZS       & 0.24                  & 0.11                  & 0.12                   & 0.09                  & 0.03                  & 0.04                   & 0.25                  & 0.10                  & 0.12                   & -0.10                 & 0.03                  & -0.05                  & -4.9                                           \\
Vanilla FT       & 0.41                  & 0.40                  & 0.39                   & 0.23                  & 0.22                  & 0.22                   & 0.40                  & 0.39                  & 0.38                   & 0.51                  & 0.43                  & 0.47                   & -3.6                                           \\
Control Codes FT & 0.43                  & 0.41                  & 0.40                   & 0.25                  & 0.24                  & 0.24                   & 0.41                  & 0.39                  & 0.39                   & 0.51                  & 0.43                  & 0.47                   & -3.6                                           \\ \bottomrule
\end{tabular}
\caption{Full fine-tuning results for T5-Large: precision, recall, and F1-scores for ROUGE-1, ROUGE-2, ROUGE-L, and BERTScore; BARTScore.}
\label{tab:full_finetuning}
\end{table}
\end{landscape}

\begin{table*}[]
\begin{tabular}{>{\raggedright}p{7.5cm}>{\raggedright}p{7.5cm}}
\toprule
News On the Web Answer & Questions \tabularnewline \midrule

Several human-induced factors are contributing to a spread in water conditions that lead to harmful algal blooms around the world. The first is land management. Agricultural runoff, such as fertilizer, and urban runoff, such as sewage, are depositing extra nutrients into bodies of water, stimulating eutrophication. The construction of dams has also slowed water movement in some areas, creating stagnant bodies that foster cyanobacteria. But climate change is also leading to hotter temperatures and warmer bodies of water that stimulate the growth of cyanobacteria . Droughts also lead to water and @ @ @ @ @ @ @ @ @ @ we continue to see extreme warmth, if we continue to see droughts,'' Caron said, ``we will probably see more of these toxins produced and the potential for more animals and humans to get in the way of those toxins.'' & \centerline{\textbf{Reference}} Why are blue-green algae spreading? \newline \centerline{\textbf{Fine-Tuned}} \textbf{Vanilla} What causes algal blooms?' \newline
\textbf{Control Codes} What causes these algal blooms?' \newline 
\centerline{\textbf{Zero-Shot}}
\textbf{T5-Small} sewage, are depositing extra nutrients into bodies of water, stimulating eutrophication' \newline
\textbf{BART-Large} Several human-induced factors are contributing to a spread in water conditions that lead to harmful algal blooms around the world. The first is land management. Agricultural runoff, such as fertilizer, and urban runoff, which are both increasing the amount of nutrients in the water, are depositing extra nutrients into bodies of water, stimulating eutrophication. The construction of dams has also slowed water movement in some areas, creating stagnant bodies that foster cyanobacteria. But climate change is also leading to hotter temperatures and warmer bodies of food that stimulate the growth of cyanob bacteria. Droughts also lead to water pollution. ``If we continue to see extreme heat and @ @ @ \& @ @ \# @ @ we continue- see extreme warmth, if we continue, see droughts, '' Caron said, `` we will probably see more of these toxins produced and the potential for more animals and humans to get in the way of those toxins. ''\thinspace' \newline
\textbf{T5-Base} warmer bodies of water that stimulate the growth of cyanobacteria' \newline
\textbf{T5-Large} harmful algal blooms' \newline
\textbf{GPT-2} 58 8 8' \newline
\textbf{T5-3B} Agricultural runoff, such as fertilizer, and urban runoff, such as sewage' \newline
\textbf{T0-3B} What is the reason for the spread of harmful algal blooms?' \newline
\textbf{T0pp} What is the main cause of cyanobacteria?'
\tabularnewline
\bottomrule
\end{tabular}
\caption{Examples from the fine-tuning dataset (News On the Web).}
\label{tab:now_examples}
\end{table*}


\begin{table*}[]
\begin{tabular}{>{\raggedright}p{7.5cm}>{\raggedright}p{7.5cm}}
\toprule
Multi-News Summary & Generated Questions \tabularnewline \midrule

jpmorgan chase ceo jamie dimon plans to apologize before members of congress tomorrow for a trading loss that has cost the bank more than \$ 2 billion. he also will say that the bank has taken steps to make sure it does not happen again. ``we have let a lot of people down, and we are sorry for it,'' dimon says in testimony prepared for his appearance before the senate banking committee. ``while we can never say we won't make mistakes—in fact, we know we will—we do believe this to be an isolated event,'' dimon says. the wall street journal has the full text, in which dimon also castigates traders for badly misunderstanding the risks involved and ``generally ineffective'' managers for failing to rein them in, notes the journal. 
& \centerline{\textbf{Fine-Tuned}} 
\textbf{Vanilla} What will jpmorgan chase ceo jamie dimon say?' \newline 
\textbf{Control Codes} What will the ceo say?' \newline
\centerline{\textbf{Zero-Shot}}
\textbf{T5-Small} jpmorgan chase ceo jamie dimon' \newline 
\textbf{BART-Large} jpmorgan chase ceo to apologize for trading lossJPMorgan Chase ceo jamie dimon plans to apologize before members of congress tomorrow for a trading loss that has cost the bank more than \$ 2 billion. he also will say that the bank has taken steps to make sure it does not happen again. `` we have let a lot of people down, and we are sorry for it, '' dimon says in testimony prepared for his appearance before the senate banking committee. `` while we can never say we won\'t make mistakes — in fact, we know we will — we do believe this to be an isolated event, ''dimon says. the wall street journal has the full text, in which dimon also castigates traders for badly misunderstanding the risks involved and `` generally ineffective '' managers for failing to rein them in, notes the journal.' \newline
\textbf{T5-Base} trading loss' \newline
\textbf{T5-Large} senate banking committee' \newline
\textbf{GPT-2} 18 18 18’ \newline
\textbf{T5-3B} jpmorgan chase ceo jamie dimon plans to apologize before members of congress tomorrow for a trading loss that has cost the bank more than \$ 2 billion' \newline
\textbf{T0-3B} dimon to apologize before congress for trading loss' \newline
\textbf{T0pp} jpmorgan chief to apologize to congress for trading loss' \tabularnewline
\bottomrule
\end{tabular}
\caption{Examples from the Multi-News dataset.}
\label{tab:mn_examples}
\end{table*}


\begin{table*}[]
\centering
\begin{tabular}{>{\raggedright}p{15cm}}

\toprule

\textbf{Multi-News Summary} some 33 years after it was launched in 1977, voyager 1 has reached the outer edge of the solar system and is on course to become the first man-made device to sail into the vast stretches of space that lie beyond. astronomers have confirmed that the spacecraft has reached a region called the heliopause, where the solar winds that have blown past voyager for the last 10 billion miles slow to a stop, discover reports. in another few years, the spacecraft will emerge from the shell of gases that surrounds the solar system and enter interstellar space. ``when voyager was launched, the space age itself was only 20 years old, so there was no basis to know that spacecraft could last so long,'' a project scientists tells the bbc. ``we had no idea how far we would have to travel to get outside the solar system. we now know that in roughly five years, we should be outside for the first time.''
 \newline \textbf{NewsQs Question} When will voyager 1 reach the outer edge of the solar system?
 \newline \textbf{Annotations} acceptable, acceptable \tabularnewline
\midrule

\textbf{Multi-News Summary} the annual report from the sustainable development solutions network on the world's happiest and least happy countries is out, and if you live in the united states, sorry: the us doesn't make it into the top 10( it's ranked no. 13). the survey ranks 157 countries using factors including gdp, years of healthy life expectancy, freedom from business and government corruption, and "having someone to count on in times of trouble." the happiest: denmark switzerland iceland norway finland the least happy( in order from most to least happy): benin afghanistan togo syria burundi the sdsn notes that the editors of the list are encouraging a focus on "happiness inequality," saying that they have found such inequality has increased, and that people are happier in societies where there is more happiness equality. click for the top and bottom 10 in each category from reuters. \newline
\textbf{NewsQs Question} the report on happiness inequality has been released. which countries have improved their ranking? \newline
\textbf{Annotations} narrow topic, acceptable
\tabularnewline
\midrule

\textbf{Multi-News Summary} more than a week after america went to the polls, seven house races are still too close to call and two statewide elections are undecided. the results look certain to make the historic republican win in the house even bigger, the los angeles times notes, with democratic candidates defending all seven districts and leading in just two — california's 11th and kentucky's 6th. the differential is fewer than 700 votes in most races; a 10th race in north carolina's 2nd has been called for the republican, but a recount is expected. alaska is preparing to examine write-in ballots to decide its senate race; in minnesota, the last unresolved governor's race in the nation shows no sign of being decided anytime soon, the ap reports. republican tom emmer trails democrat mark dayton by 8,750 votes — well within the margin for a recount — and the fight is expected to continue into december. gov. tim pawlenty has held transition talks with both men and may have to extend his term if the race isn't resolved by jan. 3. \newline
\textbf{NewsQs Question} what's the difference between a republican and a democratic candidate? \tabularnewline
\textbf{Annotations} broad topic, broad topic \tabularnewline
\bottomrule
\end{tabular}
\caption{Examples of NewsQs human evaluation.}
\label{tab:human_eval_examples}
\end{table*}

\end{document}